\documentclass[10pt,twocolumn,letterpaper]{article}

\usepackage{cvpr}
\usepackage{times}
\usepackage{epsfig}
\usepackage{graphicx}
\usepackage{amsmath}
\usepackage{amssymb}
\usepackage{algorithm}
\usepackage{algpseudocode}
\usepackage{mathtools}

\usepackage[utf8]{inputenc}
\usepackage{fourier} 
\usepackage{array}
\usepackage{makecell}
\usepackage{multirow}

\DeclarePairedDelimiter\floor{\lfloor}{\rfloor}

\usepackage{tabularx}
\usepackage{booktabs}
\usepackage{enumitem}

\usepackage[pagebackref=true,breaklinks=true,letterpaper=true,colorlinks,bookmarks=false]{hyperref}

\cvprfinalcopy 

\def\cvprPaperID{****} 

\ifcvprfinal\fi
\begin{document}

\title{Variable-Viewpoint Representations for 3D Object Recognition}

\author{
Tengyu Ma, Joel Michelson, James Ainooson, Deepayan Sanyal,\\
Xiaohan Wang, Maithilee Kunda \\
\textit{Department of Electrical Engineering and Computer Science, Vanderbilt University} \\
\{tengyu.ma, joel.p.michelson, james.ainooson, deepayan.sanyal,\\
xiaohan.wang, mkunda\}@vanderbilt.edu
}

\maketitle

\begin{abstract}
For the problem of 3D object recognition, researchers using deep learning methods have developed several very different input representations, including ``multi-view'' snapshots taken from discrete viewpoints around an object, as well as ``spherical'' representations consisting of a dense map of essentially ray-traced samples of the object from all directions.  These representations offer trade-offs in terms of what object information is captured and to what degree of detail it is captured, but it is not clear how to measure these information trade-offs since the two types of representations are so different.  We demonstrate that both types of representations in fact exist at two extremes of a common representational continuum, essentially choosing to prioritize either the number of views of an object or the pixels (i.e., field of view) allotted per view.  We identify interesting intermediate representations that lie at points in between these two extremes, and we show, through systematic empirical experiments, how accuracy varies along this continuum as a function of input information as well as the particular deep learning architecture that is used.
\end{abstract}


\section{Introduction}
\label{sec:intro}

\noindent Imagine you are visiting a museum to photograph a historical artifact.  Because of the limited disk size of your camera, you can only store a total of $1920\times1080$ pixels. Every view of this artifact is appealing, and so you face a difficult choice: will you take a single high resolution picture ($1\times1080p$) of the front-view only, several medium resolution pictures ($6\times480p$) of a few more views, or a panoramic picture with as many views as possible?  Which approach would help you remember the artifact best?  

This question has emerged as a key point of differentiation among current, high-performing approaches to 3D object recognition.  Multi-view representations~\cite{b:mvcnn, b:rotationnet} essentially capture a discrete collection of whole-object views from various viewpoints.  Spherical representations~\cite{b:s2cnn, b:spcnn, b:ugscnn} capture a continuous sampling of single-point views from all around the object.  (We focus here on approaches to 3D object recognition using inputs that are projections of 3D information onto 2D image arrays.  Other approaches use inputs represented explicitly in 3D, such as voxels~\cite{b:voxnet, b:3DShapeNets_ModelNet} and point-clouds~\cite{b:pointlib, b:pointnet}, but these fall outside the scope of our discussions in this paper.)

\begin{figure}[t]
\begin{center}
   \includegraphics[width=\linewidth]{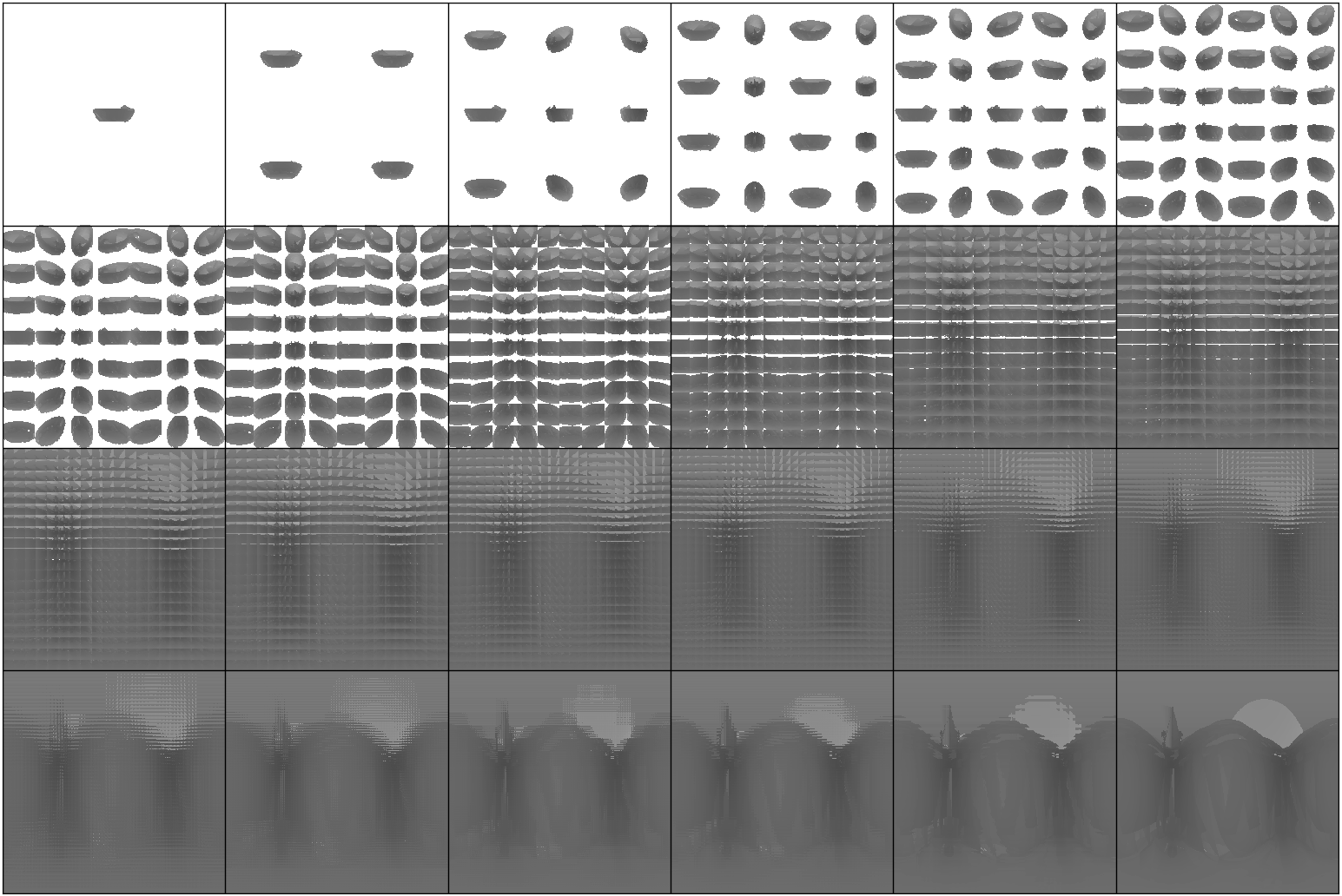}
\end{center}
   \caption{The continuum of our Variable Viewpoint ($V^2$) representations from the Multi-View extreme (top left) to the Spherical extreme (bottom right). All representations are from the same 3D object: a bathtub in ModelNet40.}
\label{fig:p-s_overview}
\end{figure}

Which representation is better, multi-view or spherical?  Both yield impressive results with various deep learning architectures, but it is difficult to make generalizable, apples-to-apples judgments between them when they draw from such different types of input information.  
\begin{itemize}[nolistsep,noitemsep]
    \item In this paper, we reveal that multi-view and spherical representations are essentially two extreme cases of a unified representational continuum, which we call the Variable Viewpoint ($V^2$) representation, and which is illustrated in Figure~\ref{fig:p-s_overview}.
    \item We show, through systematic empirical experiments, how 3D object recognition accuracy varies along this $V^2$ continuum as a function of the input information available to the learner, and we find an interesting ``S'' shaped pattern that seems to exhibit two broad regions of high-performing representations within this continuous $V^2$ input space.
    \item We also demonstrate that some learning architectures perform well only in certain regions of the $V^2$ input space, while others show robust performance across the entire space.
\end{itemize}

\section{Variable Viewpoint Representations}
\label{sec:method}

\begin{figure}[b]
\begin{center}
   \includegraphics[width=\linewidth]{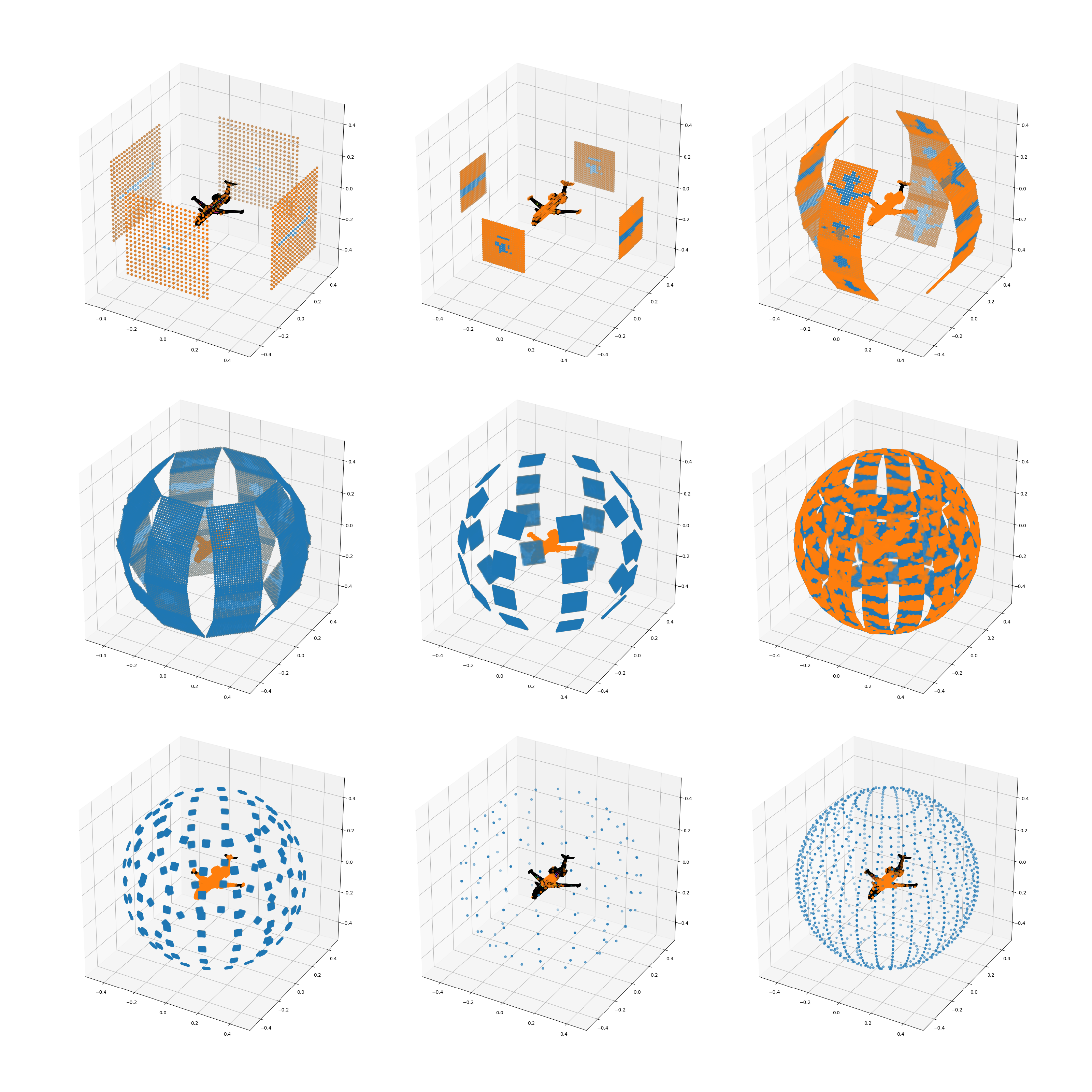}
\end{center}
   \caption{Samples of $V^2$-generating configurations. Each blue dot is a sampling point shooting a ray to the object, and each orange dot is the intersection point between the ray and the object surface. The object is a normalized airplane in ModelNet40.}
\label{fig:cam_config}
\end{figure}

\noindent We define \textit{variable viewpoint ($V^2$)} representations as a collection of samples of 2D projections from around a 3D object, specified by following parameters:
\begin{enumerate}[noitemsep,leftmargin=*]
    \item The \textbf{number} of views (e.g., akin to the number of cameras positioned around an object)
    \item The \textbf{position} of views (where each camera is located)
    \item The \textbf{size} of each view (field of view of each camera)
    \item The \textbf{density} of each view (resolution of each camera)
\end{enumerate}

The number and the position determine the view distribution, and the size and the density determine the view resolution. If the view distribution is sparse and the resolution is high, $V^2$ is equivalent to multi-view representations.  If the distribution is dense and the resolution is only one pixel, $V^2$ is equivalent to spherical representations.

$V^2$ representations are generated by first centering and normalizing a given 3D object inside a sphere.  
Then, each sampling point on the sphere becomes the center point of a view plane that is tangent to the sphere.  
On each view plane, sampling points form a evenly spaced grid surrounding the center point. All points on a plane will shoot rays towards the object that are perpendicular to the plane, finally reaching the object (or in some cases missing it entirely, becoming a ``background'' pixel in the resulting projection).  Figure~\ref{fig:cam_config} shows intuitive examples of different generating configurations of $V^2$.

The information obtained from each ray when it intersects the object populates the channels of the 2D projection.  For example, this information could consist of 
the travelling distance, the RGB value on the surface, or the incident angle.

\subsection{Number of Views vs Pixels per View}

\begin{figure}[b]
\begin{center}
   \includegraphics[width=\linewidth]{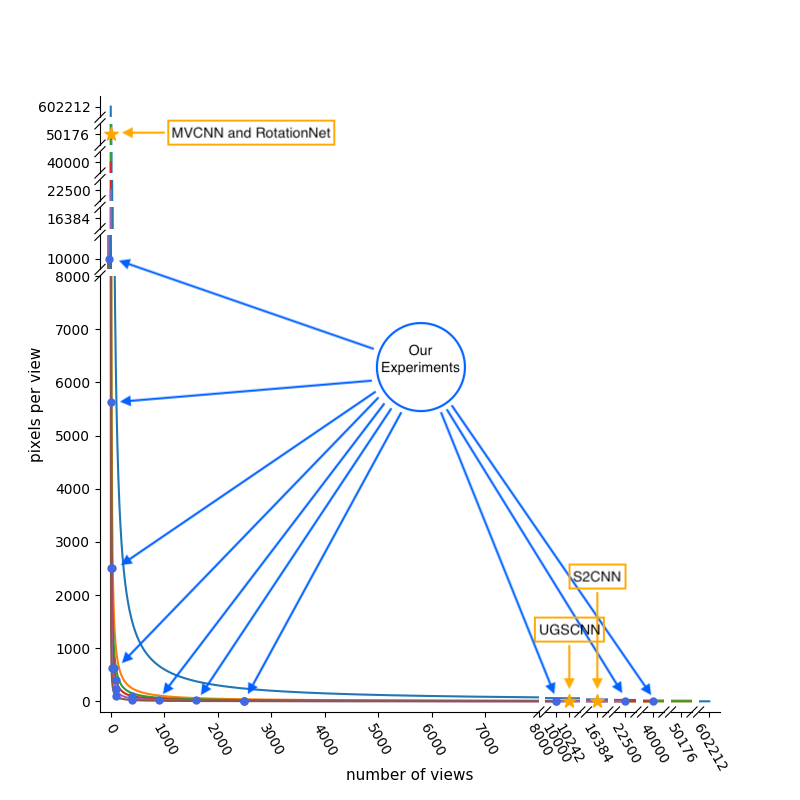}
\end{center}
   \caption{By fixing the total number of pixels $C$, the number of views $NV$ (x-axis) and pixels per view $PV$ (y-axis) will form a reciprocal curve. Different curves (a bit hard to see here) represent different values for the total number of pixels $C$, with fewer-pixel curves being closer to the origin. Orange stars indicate where multi-view \cite{b:mvcnn,b:rotationnet} and spherical representations \cite{b:s2cnn,b:ugscnn} from existing research exist in this space.  Notice the rich region in the middle waiting to be explored.  Our experiments in this paper examine $V^2$ representations at the locations of the blue dots.}
\label{fig:nvpv}
\end{figure}

On the principle that more information is better, we might expect the best $V^2$ representations to be those that use lots of cameras, positioned all around an object, where each camera has a super-wide field of view and a super-high resolution.  But, which of these parameters is the most helpful to maximize?

Of the many tradeoffs that could be considered, we focus here on the tradeoff between the number of views versus the pixels per view.  As in our museum example at the beginning, suppose we control for a constant amount of input information by holding fixed the total number of pixels available.  We could either spend those pixels on a small number of views, with many pixels per view (i.e., multi-view approach), or on a large number of views, with one pixel per view (i.e., spherical approach), or something in the middle.

Figure~\ref{fig:nvpv} shows how $V^2$ representations with a fixed number of total pixels fall along reciprocal curves in the space defined by number of views versus pixels per view.  In this figure, the orange stars show where existing work on multi-view (top left) and spherical representations (bottom right) would be located.  There is a rich region of representations in the middle are waiting to be explored.

\subsection{$V^2$ specification for this paper}
\label{sec:v2}

There would be many ways to define numerical parameters to capture the $V^2$ framework.  In this paper, we define five specific parameters:
\begin{enumerate}[nolistsep,noitemsep]
    \item $m$ = number of rows in view distribution (e.g. latitude lines)
    \item $n$ = number of columns in view distribution (e.g. longitude lines)
    \item $x$ = width of each view, in pixels
    \item $y$ = height of each view, in pixels
    \item $d$ = density of pixels in a view, i.e., distance between sampling rays in a view place
\end{enumerate}

One additional parameter, the number of channels $NC$, specifies what type of information is captured by each sampling ray.

If we denote the Number of Views as $NV$, Pixels per View as $PV$, and total pixels as $C$,
the following equations hold:
$m\times n = NV$, $x\times y = PV$, and $C = NV \times PV \times NC$.

\begin{figure}[t]
\begin{center}
   \includegraphics[width=\linewidth]{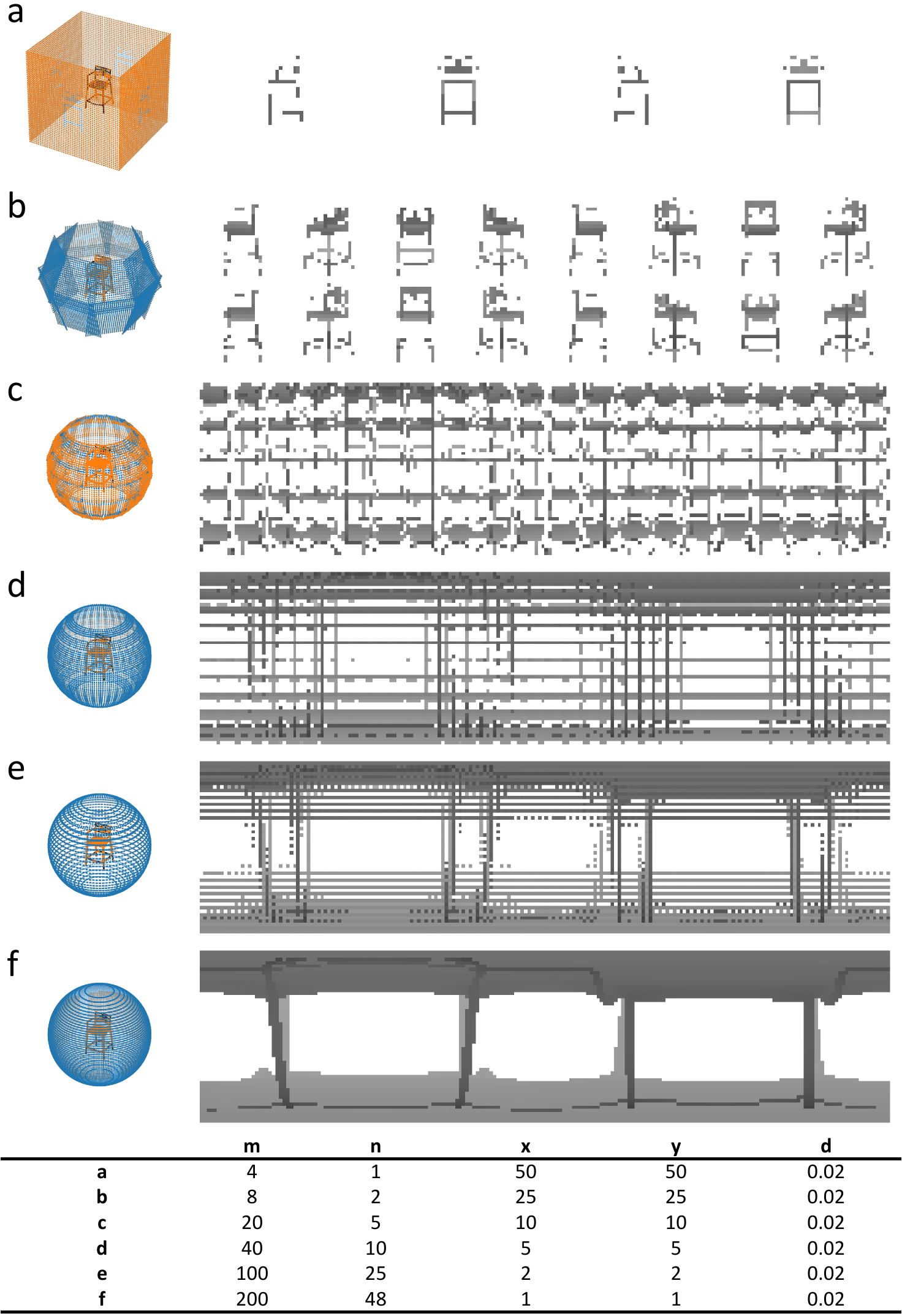}
\end{center}
   \caption{The evolution of $V^2$ from Multi-View to Sphere using the same number of total pixels. The left shows the sampling point distribution, the right shows the generated $V^2$ representation, and the bottom table shows the five parameters for each.}
\label{fig:evolving}
\end{figure}

Next, we describe in detail how we generate $V^2$ representations using these parameters.  Figure~\ref{fig:evolving} shows a concrete example of the range of $V^2$ representations from Multi-View~\cite{b:mvcnn} to Sphere~\cite{b:s2cnn}, while keeping the total number of pixels constant.

\begin{table*}[t]
\label{tb:comparison}
\begin{center}
\setlength\tabcolsep{5pt}
\renewcommand{\arraystretch}{0} 
\begin{tabular}
{@{} l c c c c c c c c c c @{}}
\hline
\multirow{2}{*}[1em]{\thead[l]{}}
 & \multicolumn{9}{c}{\textbf{$V^2$ Representation}} & \multirow{2}{*}[1em]{\thead{\textbf{Network} \\ \textbf{Architecture}}} \\
 & \thead{$m$}
 & \thead{$n$}
 & \thead{$x$}
 & \thead{$y$}
 & \thead{$d$}
 & \thead{$NV$}
 & \thead{$PV$}
 & \thead{$NC$}
 & \thead{$C$}
 & \\
\hline
\thead[l]{\textbf{Su \textit{et al.}}~\cite{b:mvcnn} $^1$}
& 12 & 1 & 224 & 224 & n/a
& 12 & 50176 & 1 & 602212
& MVCNN \\
\thead[l]{\textbf{Kanezaki \textit{et al.}}~\cite{b:rotationnet} $^2$}
& 12 & 1 & 224 & 224 & n/a
& 12 & 50176 & 1 & 602212
& RotationNet \\
\thead[l]{\textbf{Taco \textit{et al.}}~\cite{b:s2cnn} $^3$}
& \multicolumn{2}{c}{16384} & 1 & 1 & n/a
& 16384 & 1 & 6 & 98304
& S2CNN  \\
\thead[l]{\textbf{Chiyu \textit{et al.}}~\cite{b:ugscnn} $^4$}
& \multicolumn{2}{c}{10242} & 1 & 1 & n/a
& 10242 & 1 & 6 & 61452
& UGSCNN \\
\hline
\multirow{3}{*}[-5em]{\textbf{This paper}}
    & \thead{4 \\ 8 \\ $\vdots$ \\ 100}
    & \thead{1 \\ 2 \\ $\vdots$ \\ 25}
    & \thead{50 \\ 25 \\ $\vdots$ \\ 1}
    & \thead{50 \\ 25 \\ $\vdots$ \\ 1}
    & $\dfrac{1}{50}$
    & \thead{4 \\ 16 \\ $\vdots$ \\ 10000}
    & \thead{2500 \\ 625 \\ $\vdots$ \\ 1}
    & 1
    & 10000
    & \thead{MVCNN \\ RotationNet \\ S2CNN \\ UGSCNN \\ ResNet} \\
\cline{2-11}
    & \thead{4 \\ 12 \\ $\vdots$ \\ 300}
    & \thead{1 \\ 3 \\ $\vdots$ \\ 75}
    & \thead{75 \\ 25 \\ $\vdots$ \\ 1}
    & \thead{75 \\ 25 \\ $\vdots$ \\ 1}
    & $\dfrac{1}{75}$
    & \thead{4 \\ 36 \\ $\vdots$ \\ 22500}
    & \thead{5625 \\ 625 \\ $\vdots$ \\ 1}
    & 1
    & 22500
    & \thead{MVCNN \\ RotationNet \\ S2CNN \\ UGSCNN \\ ResNet} \\
\cline{2-11}
    & \thead{4 \\ 8 \\ $\vdots$ \\ 400}
    & \thead{1 \\ 2 \\ $\vdots$ \\ 100}
    & \thead{100 \\ 50 \\ $\vdots$ \\ 1}
    & \thead{100 \\ 50 \\ $\vdots$ \\ 1}
    & $\dfrac{1}{100}$
    & \thead{4 \\ 16 \\ $\vdots$ \\ 40000}
    & \thead{10000 \\ 2500 \\ $\vdots$ \\ 1}
    & 1
    & 40000
    & \thead{MVCNN \\ RotationNet \\ S2CNN \\ UGSCNN \\ ResNet} \\
\hline              
\end{tabular}
\end{center}
\caption{Using the parameters described in Section~\ref{sec:v2}, we characterize the multi-view and spherical representations from four previous research studies. 
$^1$ $^2$ Each view has a single depth channel. $^3$ $^4$ Adopted different sphere sampling methods whose $m$ and $n$ are replaced by the number of sphere sampling points.
$^3$ $^4$ Each view has 6 channels consisting of the depth, the $\cos/\sin$ of incident angle of the object and its convex hull.}
\end{table*}

\textbf{Data Normalization.}  
The containing sphere is defined with diameter of 1 so the ray travelling distance will range from $[0.0, 1.0]$. A given 3D object will be normalized such that the longest side of its bounding box will be scaled to 0.4 while keeping the same aspect ratio, and the geometric center of the bounding cuboid is regarded as the center of this object.

This normalizing process guarantees that the normalized object will be completely contained inside the sphere, since the worst cast is that the bounding box has size $0.4\times0.4\times0.4$, whose circumscribed sphere having diameter of $2\times0.2\times\sqrt{3}\approx0.7$ leaves enough space for the containing sphere.

\textbf{View Center.}  
View centers are sampling points on the surface of the sphere.
The distribution of these points is its own research problem, sphere sampling~\cite{b:sphere_sampling_survey}.  Here, we adopt a straightforward approach for generating regular grid points on the sphere along lines of longitude and latitude.
In polar coordinates, $\theta, \phi$ is the azimuthal and polar angle respectively.
$\theta$ is evenly sampled $m$ times from $[0, 2\pi)$, and $\phi$ is evenly sampled $n$ times from $(0, 1)$, updated by $\phi=\arccos{(1-2\phi)}$ to avoid polar clustering.

\textbf{View Grid.}  
\label{sec:view_plane_grid}
Each view plane is a tangent plane contacting the sphere at each view center.
Plane sampling~\cite{b:plane_sampling} is also another research area, we slightly refined the approach of centric systematic~\cite{b:centric_sampling} sampling to generate the view grid.
Every point on the grid is symmetrically surrounding the center point, constructing a matrix of dimension $x$ by $y$ and shooting a ray parallel to its plane normal. In particular, the position of the center point is $(\floor*{x / 2}, \floor*{y / 2})$.
The closest point interval $d$ determines the density of each view plane, and we set $d=1/x$ to guarantee all rays will intersect with the object.

\section{Experiments}

Table~\ref{tb:comparison} shows how previous research can fit into our $V^2$ framework and how our experiments examine more general combinations of parameters.




We conducted our experiments to investigate the recognition performance and architecture preference on many combinations of pre-defined $V^2$ representations, ranging from less views to more views, as shown in Table~\ref{tb:comparison}, and also covering various deep architectures, including MVCNN~\cite{b:mvcnn} and RotationNet~\cite{b:rotationnet} (designed for multi-view), S2CNN and UGSCNN~\cite{b:ugscnn} (designed for sphere), and ResNet18~\cite{b:resnet} as a baseline.
All hyper-parameters follow the default values reported in each paper for approaching their best performance.
We used two datasets: ModelNet40~\cite{b:3DShapeNets_ModelNet} and SHREC17~\cite{b:shrec17}, a competition version of ShapeNet~\cite{b:shapenet}.

\subsection{$V^2$ Configurations}
\label{sec:data_conf}
The total number of pixels, $C$, has been set to 10000, 22500, and 40000, as they equal to $1\times4\times25\times25$, $1\times4\times50\times50$, $1\times4\times75\times75$, and $1\times4\times100\times100$ indicating the initial configuration.
Referring to the $V^2$ representation section, $1\times4\times100\times100$ means that $m=1, n=4, x=100, y=100$ yields a representation of 4 views and 10000 pixels per view. This is a starting point at the Multi-View end.

Intermediate-level $V^2$ representations are configured by identically subdividing each view into 4 quadrants, so $1\times4\times100\times100$ will be decomposed to $2\times8\times50\times50$. This process will keep constructing new representations until reaching the Sphere end, i.e., $100\times400\times1\times1$.

\subsection{Network Architectures}
Two networks MVCNN~\cite{b:mvcnn} and RotationNet~\cite{b:rotationnet} were involved to represent the architectures particularly designed for Multi-View, and two other networks S2CNN~\cite{b:s2cnn} and UGSCNN~\cite{b:ugscnn} were involved to represent the architectures particularly designed for Spherical representations.
ResNet18~\cite{b:resnet} is adopted as a baseline.

The main difference among these architectures is the convolutional pattern that they use.
ResNet is the classic architecture, that is, standard 2D convolutional kernels will travel around the whole image, with no view splitting.
MVCNN and RotationNet assembled replications of a classic architecture; if the whole image contains many views, each view will have its own convolutional kernels travelling around, i.e., view splitted.
S2CNN and UGSCNN use specially designed spherical kernels to convolve/cross-correlate the whole input in the $S^2$ or $SO(3)$ space, and there is no view splitting here either.

In our experiments, we use ResNet18 as the component architecture for both MVCNN and RotationNet, and also for our baseline.  Then, for each of these three ResNet-related architectures, we experiment with two ResNet versions, pre-trained on ImageNet~\cite{b:imagenet} or not.

The first convolutional layers of these nets were slightly modified to handle different input sizes, but the rest of layers were all kept the same.
Every network architecture was expected to be trained and tested on all $V^2$ representations stated in section~\ref{sec:data_conf} except MVCNN and RotationNet, whose architecture size will explode with the increasing number of views, since there is an entire separate network for each view. Thus, we limited the number of views to be less than 100 for MVCNN and RotationNet experiments.




\section{Results and Discussion}

\begin{figure}[b]
\begin{center}
   \includegraphics[width=\linewidth]{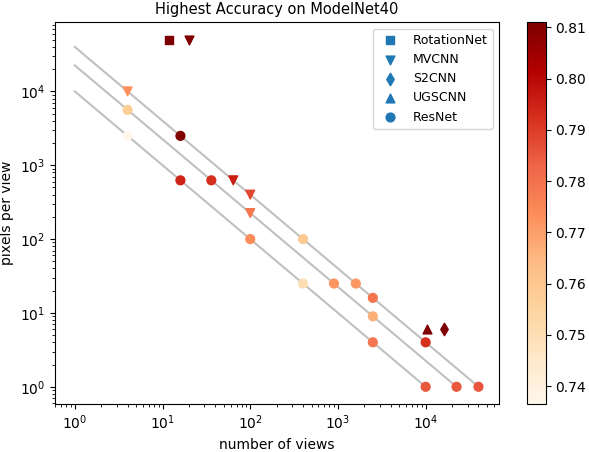}
\end{center}
   \caption{Highest accuracy across all models on ModelNet40. Points not on the lines are results from previous existing research. Axes are in a loglog scale for better visibility. Colors represent the recognition accuracy, as indicated by the color bar. Shapes represent the choice of architecture. Silver lines from left to right indicate total number of pixels $10000$, $22500$, and $40000$.}
\label{fig:highest_acc}
\end{figure}

We ran each representation-architecture pairing five times, with different random weight initializations.  All results here represent the average accuracies obtained across these five runs.  With 5 architectures, 2 datasets, and 21 configurations of $V^2$, we ended up with around 210 accuracy data points.

Figure \ref{fig:highest_acc} shows one view of our results.  We projected accuracies onto the pixels per view vs. number of views space, as in Figure \ref{fig:nvpv}, and plotted them on a log-log plot to increase clarity.  For each $V^2$ configuration, we show here the best average accuracy (color), plus the architecture that yielded best average accuracy (shape).  

Several points are noteworthy about this chart.  First, the ResNet network overall seems to do better than the other architectures, even though it was not specifically designed for 3D recognition at all.  There seems to be a bimodal distribution of maximum accuracies, with one peak over in the multi-view-like region, with around $10^1$ views, and then another peak over the in the spherical-representations-region, with around $10^4$ views.

Interestingly, the Multi-View CNN seems to benefit from moving down into the intermediate part of the $V^2$ space (top line, triangles get darker = more accurate towards the right).  Also, there seems to be a ``dead zone'' just to the right of center.  Here, even representations having fewer overall total pixels, but having either more pixels per view or more views, seem to beat representations in this ``dead zone'' even when they lie along a higher total pixel curve.

\begin{figure}[t]
\begin{center}
   \includegraphics[width=\linewidth]{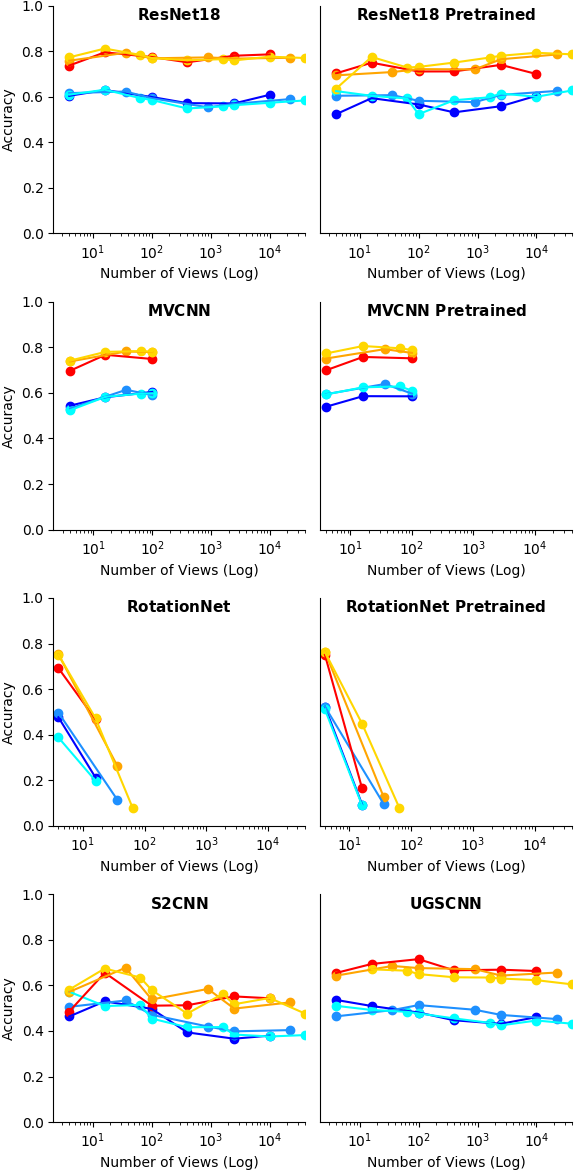}
\end{center}
   \caption{Accuracy of each network plotted as a function of the number of views. The blue series lines are the accuracy from SHREC17, and the red series lines are the accuracy from ModelNet40. The lighter the color is, more total pixels the line is for. Total pixels is in $10000$, $22500$, and $40000$.}
\label{fig:architecturelines}
\end{figure}

Figure \ref{fig:architecturelines} shows results for each individual architecture as a function of the number of views.  As is immediately apparent, RotationNet does not generalize away from multi-view representations at all.  Interestingly, the plain ResNet architecture seems the most robust in terms of generalizing across the full space of $V^2$ representations.

For both ResNet and the spherical CNN architectures, there seems to be a strong performance peak in the range of $10^1$ to $10^2$ views, and for some architectures, a more moderate rise towards the far right, when the number of views becomes very large for spherical representations.

What do these results tell us, overall?  First, there are some non-obvious patterns hidden in the middle regions of the $V^2$ space.  For instance, we might expect curves with higher absolute numbers of pixels to outperform lower curves at all points, but as Figure \ref{fig:highest_acc} shows, that is not the case.  So there seem to be some regions of $V^2$ that are providing more useful information, per pixel, than other regions.  It is an important and interesting open question to think about what is happening in these ``sweet spots'' in the $V^2$ space.

Second, not all architectures are created equal in their abilities to utilize information from across this space.  Perhaps unsurprisingly, the plain ResNet architecture shows fairly consistent performance.  More surprisingly, so does one of the spherical CNN architectures, the UGSCNN.  Also surprisingly, the plain ResNet outperforms the specialized architectures even in their ``home'' regions of the $V^2$ space, and the S2CNN actually appears to do better in the multi-view region than in its home region.  These interactions are quite complex, and more research is needed to better understand how different architectures are leveraging the input information in different ways.




\section{Related Work}
Unlike in the 2D world, raster and vector images have dominated most of the vision tasks, 3D representations keep expanding its family.
Some of them are hardly deep learning applicable such as polygon soup~\cite{b:polygon_soup, b:polygon_soup_2}, sweep-CSG~\cite{b:sweep_csg}, and spline-based representation~\cite{b:spline_based}.
Some of them need extra data processing such as polygon mesh~\cite{b:polygon_mesh}, point cloud~\cite{b:point_cloud}, and Octree/Quadtree-based representation~\cite{b:octree_quatree} in the application of Grpah CNN on Mesh~\cite{b:mesh_cnn}, PointNet~\cite{b:pointnet}, and Octree-based O-CNN~\cite{b:ocnn}.
Multi-View and Voxel are leading others in the deep learning era, as they can be directly applied to CNNs such as MVCNN~\cite{b:mvcnn} and VoxNet~\cite{b:voxnet}.
Sphere was initially proposed in 2002~\cite{b:sphere_org}, and because of its rotation-invariance property, spherical deep learning models~\cite{b:s2cnn,b:spcnn,b:ugscnn} have revealed its expressive power in 3D recognition tasks.
Because of the deep learning ease, Multi-View, Voxel, and Sphere have over-performed others to achieve the state-of-the-art accuracy on benchmark dataset, ModelNet~\cite{b:3DShapeNets_ModelNet} or ShapeNet~\cite{b:shapenet}.
Some comparisons have been done between Multi-View and Voxel~\cite{b:mv_vs_voxel}, and our work will extend these comparisons to Multi-View and Sphere by filling the transitioning gap between them, contributing an unified representation to this family.

Essentially, $V^2$ is a 2D mapping strategy; however, comparing to traditional 2D mapping algorithms~\cite{b:lscm_2dmapping, b:slim_2dmapping}, $V^2$ is easier and more regular for deep learning.
$V^2$ can support diverse distribution of viewpoints on a sphere, which will open another research question~\cite{b:sphere_sampling_survey}, such as Driscoll-Healy sampling~\cite{b:driscoll_healy_sampling}, McEwen-Wiaux sampling~\cite{b:mcewen_wiaux_sampling}, Geodesic~\cite{b:geodesic_rep}, and Goldberg. We planned to investigate the viewpoint effect particularly in the future, so aligning the purpose of this paper, we controlled this variable to a simple distribution, i.e., evenly spaced grids along longitude and latitude.

Expressive power research has been well discussed from the deep architecture perspective~\cite{b:exprpower_deep, b:exprpower_deepnet, b:exprpower_deepnet2}, 
and our work focuses on the data perspective. Given various powerful deep learning models, how could the raw input data holding enough potential features to be extracted to support those deep models?
We see our work is at the initial step of the representation learning~\cite{b:rep_learning} in the 3D recognition domain, since a 2D sampling is the very first step to extract features from a 3D ground truth. We typically skip this step in the 2D world, since most of the 2D vision data are well organized already.

However, for a specific vision task, the input representation and deep architecture together form a representing system,
and the resonance between them is important to fully utilize the power on both sides.
The smooth representation transition of $V^2$, which bridges two distinct representations before, will support us to look for the pattern of the resonance.

\section{Future Work}
In the future, we are expecting to expand the total pixels $C$ of $V^2$ to a larger number, at least comparable to the magnitude of previous 3D recognition research, for better performance comparisons for representations in the middle stage.
More channels will also be included, as besides the travelling distance of a ray from a sampling point, many other details can be captured, such as the surface color or the incident angle.
Once the middle-stage representations can reach the same level of total pixels and channels as the Multi-View and Sphere, the expressive power of them may be further explored.

Many architectures have been designed for particular representations to better utilize different structures of input representations, and thus we are expecting to design one particularly for those middle-stage representations in the future.

{\small
\bibliographystyle{ieee_fullname}
\bibliography{ref}
}

\end{document}